\documentclass[10pt, conference, compsocconf]{IEEEtran}
\usepackage{cite}
\ifCLASSINFOpdf
\else
\fi
\usepackage{amssymb}
\usepackage{graphicx}
\usepackage[turnoff]{mnotes}

\usepackage{wrapfig}
\usepackage{subfigure}

\usepackage{algorithm}
\usepackage{algorithmic}
\usepackage{color}
\usepackage{amssymb}
\usepackage{amsmath}

\newcommand{\vect}[1]{\boldsymbol{#1}}

\newcommand{\mkl}{{\em MKL}}
\newcommand{\mklsvm}{{\em $MKL_{\gamma}$}}
\newcommand{\svm}{{\em SVM}}

\newcommand{\emkl}{{\em $\epsilon$-MKL}}

\newcommand{\conc}{}
\newcommand{\esvm}{{\em $\epsilon$-SVM}}

\newcommand{\svmmetric}{{\em $SVM_{m}$}}
\newcommand{\mklmetric}{{\em $MKL_{m}$}}

\newcommand{\fda}{{\em FDA}}
\newcommand{\lmnn}{{\em LMNN}}
\newcommand{\svmfda}{{\em SVM-FDA}}

\begin{document}
%
% paper title
% can use linebreaks \\ within to get better formatting as desired
\title{A Metric-learning based framework for Support Vector Machines and Multiple Kernel Learning}

% author names and affiliations
% use a multiple column layout for up to two different
% affiliations

% conference papers do not typically use \thanks and this command
% is locked out in conference mode. If really needed, such as for
% the acknowledgment of grants, issue a \IEEEoverridecommandlockouts
% after \documentclass

% for over three affiliations, or if they all won't fit within the width
% of the page, use this alternative format:
%
\author{\IEEEauthorblockN{Huyen Do\IEEEauthorrefmark{1},
Alexandros Kalousis\IEEEauthorrefmark{2}}
\IEEEauthorblockA{\IEEEauthorrefmark{1}Computer Science Department,
University of Geneva,
Switzerland\\ Email: huyen.doth@gmail.com}
\IEEEauthorblockA{\IEEEauthorrefmark{2}Business Informatics, University of Applied Sciences Western Switzerland\\
Email: Alexandros.Kalousis@hesge.ch}
\IEEEauthorblockA{\IEEEauthorrefmark{3}\\
}}

% use for special paper notices
%\IEEEspecialpapernotice{(Invited Paper)}

% make the title area
\maketitle

\begin{abstract}
Most metric learning algorithms, as well as Fisher's Discriminant Analysis (\fda),
optimize some cost function of different measures of within-and between-class distances.
%are based on the principle of optimizing some measure of the within- and between-class distances.
On the other hand, Support Vector Machines(\svm s) and several Multiple Kernel Learning (\mkl) algorithms are based on the
\svm\ large margin theory.
Recently, \svm s have been analyzed from a metric learning perspective, and formulated as a Mahalanobis metric learning problem. This new perspective allows us to combine ideas from both \svm\ and metric learning, and to develop new algorithms that build on the strengths of each.
%Inspired by the metric learning principle, in this paper, we extend the idea to develop a metric-learning-based \svm\ framework where we incorporate the metric learning ideas into \svm\ using several different within-class distances measures.
%We further extend the idea to \mkl\ and show that \mkl\ can be also formulated as a Mahalanobis metric learning problem. We develop a family of \svm/\mkl\ methods that combine both \svm/\mkl\ and metric learning.
Inspired by the metric learning interpretation of \svm, we develop here a new metric-learning based \svm\ framework in which we incorporate metric learning concepts within \svm. We extend the optimization problem of \svm\ to include some measure of the within-class distance and along the way we develop a new within-class distance measure which is appropriate for \svm. In addition, we adopt the same approach for \mkl\ and show that it can be also formulated as a Mahalanobis metric learning problem.
Our end result is a number of \svm/\mkl\ algorithms that incorporate metric learning concepts. We experiment with them on a
set of benchmark datasets and observe important predictive performance improvements.
\end{abstract}


% For peer review papers, you can put extra information on the cover
% page as needed:
% \ifCLASSOPTIONpeerreview
% \begin{center} \bfseries EDICS Category: 3-BBND \end{center}
% \fi
%
% For peerreview papers, this IEEEtran command inserts a page break and
% creates the second title. It will be ignored for other modes.
\IEEEpeerreviewmaketitle

\section{Introduction}
\label{Introduction}
Support Vector Machines \cite{Taylor2000} have been an active research area for more than two decades. They have been widely used not only because
of their excellent predictive performance but also because their generalization ability is supported by solid generalization error bounds
defined over the radius-margin ratio.  Mahalanobis metric learning has started attracting significant attention rather recently
\cite{Lanckriet2004,Ong2003,Sonnenburg2006a,rakotomamonjy2008,Cortes2010l2,weinberger2009distance,goldberger2005nca,globerson2006mlc}.
However, relying mostly on intuition, it still lacks theoretical support. Very recently \svm\ has been reformulated in the metric learning context
and has been shown to be equivalent to a Mahalanobis metric learning problem \cite{Huyen2011}. This new interpretation of \svm\ brings the worlds
of \svm\ and metric learning together into a single unified view. This allows us to exploit the advantages of each one to develop for example hybrid
algorithms or to derive theoretical error bounds for metric learning problems exploiting the \svm\ error bounds.

In this paper we build on the ideas presented in \cite{Huyen2011} to develop a novel metric-learning-based \svm\ framework and equip \svm\ with a
metric learning bias. More precisely we will define new \svm\ optimization problems that will make use of both the between-and within-class distances.
Under the metric learning view of \svm, the margin plays the role of the between-class distance. However, \svm\ ignores the within-class
distance. In the metric-learning-based \svm\ framework that we present here, we maximize the \svm\ margin and minimize some measure of the
within-class distance. We can use different measures of the within-class distance with \svm\ and we will define a new such measure that is
more appropriate for \svm. We will give a new \svm\ algorithm that optimizes both the margin and the new within-class distance measure
that we define. The resulting optimization problem is convex and can be directly kernelized.
Moreover we will follow the same approach with \mkl\ and show that it, also, can be formulated as a Mahalanobis metric learning problem. As
a result we develop a novel family of \mkl\ methods that incorporate the metric learning bias. We experiment with the developed algorithms
on a number of benchmark datasets and saw that the incorporation of the within-class distance measures in the \svm\ learning problem brings
significant performance improvements.

Finally, we give a unified view of \svm, metric-learning-based \svm, metric learning algorithms and Fisher Discriminant Analysis (\fda) using the
concepts of between-class and within-class distances. This view provides new insights to the existing algorithms and unveils some
unexpected relations.

The rest of the paper is organized as follows. In the next section we briefly describe the basic concepts of \svm, \fda\ and Mahalanobis metric
learning. In section \ref{related} we summarize the metric learning view of \svm. In section \ref{sec:emkl} we propose the metric-learning-based
\svm\ framework which relies on the use of the between-class and within-class distance measures. In section \ref{sec:unifiedview} we provide a
common view of \svm, metric-learning based \svm, \fda\ and metric learning algorithms.
%Section \ref{sec:MKLinMetricview} extends the metric learning perspective to Multiple Kernel Learning which forms the metric-learning-based \mkl\ framework.
In Section 5 we employ the metric learning perspective in the context of \mkl\ and develop the metric-learning-based \mkl\ framework.
We report experimental results in section \ref{sec:expr} and conclude with section \ref{sec:conclusion}.

\section{Preliminary}
Consider a binary classification problem in which we are given a set of $n$ learning instances
$S= \{({\mathbf x}_{1}, y_{1}),..., ({\mathbf x}_{n},y_{n})\}, {\mathbf x}_i \in {\mathcal R}^d$, where $y_i$ is the class label of the ${\mathbf x}_i$ instance and $y_i \in \{+1, -1\}$. Let $\mathcal D_i = \{{\mathbf x}_j| y_j = C_i\}$ be the samples of class $C_i$.
%We denote by $\|{\mathbf x}\|_p$ the $l_p$ norm of the ${\mathbf x}$ vector.
\subsection{Support Vector Machines and MKL}
% Let  $H_{\mathbf w}$ be the hyperplane given by ${\mathbf w}^T{\mathbf x} + b=0$; the signed distance, $d(\vect {x}_i, H_{\mathbf w})$, of some point $\vect {x}_i$ from the $H_{\mathbf w}$ hyperplane is:
%  $d(\vect {x}_i, H_{\mathbf w})=\frac{{\mathbf w}^T\vect{x_i} + b}{\|{\mathbf w}\|_p}$.

\svm s learn a hyperplane $H_{\mathbf w}: {\mathbf w}^T{\mathbf x} + b=0$ which maximizes the margin between the two classes. The \svm\ margin is defined informally as the distance of the nearest instances from $H_{\mathbf w}$
 \cite{Taylor2000}.
The \svm\ optimization problem is:
% \begin{eqnarray}
% \label{eq:SVM-std-margin}
% \max_{{\mathbf w},b, \gamma} \,\, \,\, \gamma,   \,\,\,\,\,\,\,
% s.t.   \,\,\,\,\       \frac{y_i({\mathbf w}^T {\mathbf x}_i + b)}{\|\mathbf w\|_2} \geq \gamma,  \ \forall i\
%  \end{eqnarray}
%  which is usually rewritten:
\begin{eqnarray}
\label{eq:SVM-std}
\min_{\mathbf{w},\vect{\xi}, b} &&  \|{\mathbf w}\|^2_2  + C \sum_{i=1}^l \xi_i \\
 s.t.  &&   y_i({\mathbf w}^T {\mathbf x}_i + b) \geq 1 - \xi_i, \forall i \nonumber
\end{eqnarray}
The performance of \svm\ strongly depends on the choice of kernel. \mkl\ addresses that problem by selecting or learning the appropriate kernel(s) for a given
problem \cite{Lanckriet2004,Cristianini2002,Ong2003,rakotomamonjy2008,Cortes2010,Sonnenburg2006a,Argyriou05}. Given a set of basis kernel functions,
$\vect{Z}=\{K_{k}({\mathbf x},{\mathbf x}')|k:=1\dots m \}$,  \mkl\ learns a kernel combination, usually convex, so that some cost function, e.g margin, kernel
alignment,  is optimized. The cost function that is most often used in \mkl\ is the margin-based objective function of \svm\ \cite{rakotomamonjy2008,Cortes2010,Sonnenburg2006a,Argyriou05,Lanckriet2004,Kloft2011}.
We denote by \mklsvm\ the \mkl\ method which learns linear kernel combinations and uses as its cost function that of standard \svm, i.e. it finds a linear kernel combination that maximizes the margin.
Its optimization problem is:
\begin{eqnarray}
\label{primalform}
\min_{{\mathbf w},b,\vect{\xi},\vect{\mu}}  &&  \frac{1}{2} \sum_{k}^m \langle {\mathbf w}_{k},{\mathbf w}_{k} \rangle + \frac{C}{2}\sum_{i=1}^{l}\xi_{i}^{2} \\ \nonumber
s.t.                             & & y_{i}(\sum_{k}^m \langle {\mathbf w}_{k},\sqrt{\mu_{k}}\mathbf \Phi_{k}({\mathbf x}_{i})\rangle + b) \geq 1-\xi_{i}, \forall i \nonumber \\
&& \|\vect{\mu}\|_p = 1, \mu_k \geq 0, \forall k  \nonumber
\end{eqnarray}
\subsection{Mahalanobis metric learning}
The squared Mahalanobis distance between two
instances ${\mathbf x}_i$ and ${\mathbf x}_j$ is:\begin{small}
%\begin{eqnarray}
%\label{eq:mahalanobis}
$d_{\mathbf M}^2({\mathbf x}_i,{\mathbf x}_j)=({\mathbf x}_i - {\mathbf x}_j)^T \mathbf M ({\mathbf x}_i - {\mathbf x}_j) $ \end{small}
 %=({\mathbf x}_i - {\mathbf x}_j)^T \mathbf A^T \mathbf A ({\mathbf x}_i - {\mathbf x}_j)
%\end{eqnarray}
where $\mathbf M$ is a positive semi-definite matrix, $\mathbf{M} \succeq 0$.
Learning a Mahalanobis metric parametrized by $\mathbf M$ ($\mathbf{M} \succeq 0$) is equivalent to learning a linear transformation $\mathbf{A}$
where $\mathbf{A}^T\mathbf{A} = \mathbf M$.

Typical metric learning methods try to bring instances of the same class close while pushing instances of different classes far away. They do so by optimizing a cost function of the Mahalanobis distance, while globally or locally satisfying some constraints on the pairwise distances. This is equivalent to minimize some
measure of the \textit{within-class distances} while maximizing the \textit{between-class distances} \cite{Huyen2011}.

A typical Mahalanobis metric learning optimization problem has the following form:
\begin{eqnarray}
 \label{metric.learning.opt}
    \min_{\mathbf{M}\succeq 0 }& F(\{d_{\mathbf M}^2({\mathbf x}_i,{\mathbf x}_j)|{\mathbf x_i, \mathbf x_j \in S}\}) &
  s.t. \,   {\mathcal Constraints}   \\
\Leftrightarrow && \nonumber \\
  \min_{\mathbf{A}} & F(\{d_{\mathbf A}^2({\mathbf x}_i,{\mathbf x}_j)|{\mathbf x_i, \mathbf x_j \in S}\}) &
  s.t.   \, {\mathcal Constraints} \nonumber
\end{eqnarray}
where $F$ and $\mathcal Constraints$ are the cost function and the constraints respectively, parametrized either by $\mathbf M$ or $\mathbf A$.
The constraints can be local, i.e applied to instance pairs that are in the same neighborhood, or global, i.e applied for all instance pairs.
They usually have the following form \cite{xing2003dml,globerson2006mlc,davis2007itm,weinberger2009distance}:
\begin{eqnarray}
 \label{metric.learning.constr}
d_{\mathbf M}^2({\mathbf x}_i, {\mathbf x}_j) &\leq & f_1,    \,if \,y_i = y_j \\
d_{\mathbf M}^2({\mathbf x}_i, {\mathbf x}_j) &\geq & f_2,   \,if \, y_i \neq y_j \nonumber
\end{eqnarray}
for some constants $f_1, f_2$.
These constraints reflect the primal bias of most metric learning algorithms, maximizing the \textit{between-class distance} and minimizing the \textit{ within-class distance}, or in other words, instances of the same class should have small distances while those of different classes should have large distances. If we take two simple measures of the within and between-class distances such as the sum of the pairwise distances of the same-class instances and the sum of the pairwise distances of the different-class instances respectively, the pairwise constraints in (\ref{metric.learning.constr}) will ensure that our between-class distance, denoted by $d_{B}$, is bigger than $f_2 \times t_1$, and our within-class distance, denoted by $d_{W}$, is smaller than $f_1\times t_2$, where $t_1, t_2$ are the numbers of instance pairs of the same and different class respectively over which we impose the constrains..
%
% We should mention that it is the first time that the interpretation of the pairwise constraints in terms of the within and between-class distance is given. Most of the metric learning algorithms can be expressed in terms of the within- and between-class distances optimization language, making the maximization of the between-class and the minimization of the within-class distances probably the most popular metric learning bias.

\subsection{Fisher Discriminant Analysis}
\fda, although usually not considered a metric learning
method, it also learns linear projections in the same manner as metric learning methods.
%in the two-class case it projects the samples onto a line with direction vector ${\mathbf w}$, projection in which the samples are well separated.
%It is equivalent to learning a diagonal transformation $\mathbf W, diag(\mathbf W) = {\mathbf w}$; therefore FDA can also be seen as a Mahalanobis metric learning algorithm.
It uses a similar learning bias as metric learning algorithms: the samples are well separated if their \textit{between-class distance} is large and their
\textit{within-class distance} is small. These quantities are defined as follows \cite{Duda2001}:
%However, unlike the constraint given in problem~(\ref{metric.distance}), FDA defines its between and within-class distances as
%follows \cite{Duda2001}:
let  the sample mean of class
$C_i$ be $\vect{m}_i = \frac{1}{n_i} \sum_{{\mathbf x} \in \mathcal D_i} {\mathbf x}$, then the within-class distance (or within-class scatter) of two classes $C_1,C_2,$ is defined as
$
s_{w} = \sum_{i = 1,2}\sum_{{\mathbf x} \in \mathcal D_i} \|{\mathbf x} - \vect{m}_i\|^2
$,
and the between-class distance (or between-class scatter) is defined as the squared distance between the means of the two classes, $s_{b} = \|\vect{m}_i - \vect{m}_j\|^2 $.
In the case of two-class problems \fda\ seeks for a projection line with a direction vector ${\mathbf w}$ which optimizes the ratio of the between-class over the
within-class distances in the projected space, its cost function is:
\begin{eqnarray}
 \max_{{\mathbf w}} \,\,\,\,J({\mathbf w}) &= & \frac{s_{b,{\mathbf w}}}{s_{w, {\mathbf w}}}
= \frac{({\mathbf w}^T (\vect{m}_1 - \vect{m}_2))^2}{\sum_i\sum_{{\mathbf x} \in \mathcal D_i}({\mathbf w}^T({\mathbf x} - \vect{m}_i))^2} \nonumber \\
 &=& \frac{{\mathbf w}^T \mathbf S_{\mathbf W}{\mathbf w}}{{\mathbf w}^T \mathbf S_{\mathbf B}{\mathbf w}}
\label{fld}
\end{eqnarray}
where $\mathbf S_{\mathbf W} = \sum_{i = 1,2}\sum_{{\mathbf x} \in \mathcal D_i} ({\mathbf x} - \vect{m}_i) ({\mathbf x} - \vect{m}_i)^T$ and
$\mathbf S_{\mathbf B} = (\vect{m}_1 - \vect{m}_2)(\vect{m}_1 - \vect{m}_2)^T$ are the within-class and between-class scatter matrices,
respectively. Another measure often used in the different variants of \fda\ is the total scatter matrix,
$\mathbf S_{\mathbf T} = \mathbf S_{\mathbf W} + \mathbf S_{\mathbf B} $, i.e
a multiply of the covariance matrix which quantifies the total data spread.
%\color{red} There is a problem here with the definition of the $S_B$ in  \cite{Duda2001} the exact definition is $\mathbf S_{\mathbf B} = (\vect{m}_i - \vect{m})(\vect{m}_i - \vect{m})^T$
%\color{black}

\section{Related work}
\label{related}
Do et al \cite{Huyen2011} recently show \svm\ can be formulated as a Mahalanobis metric learning problem in which the transformation matrix is diagonal $\mathbf W$, $diag(\mathbf W)={\mathbf w}=(w_1,\dots,w_d)^T, \|\mathbf w\|_2 = 1$. In the metric learning jargon \svm\ learns a diagonal linear transformation $\mathbf W$
and a translation $b$ which maximize the margin and place the two classes symmetrically in the two different sides of the hyperplane $H_{\vect1}:\vect{1}^T{\mathbf x}=0$. In the standard view of \svm, the space is fixed and the hyperplane is moved around to achieve the optimal margin. In the metric view of \svm, the hyperplane is fixed to $H_{\vect 1}$ and the space is scaled, $\mathbf W$, and then translated, $b$, so that the instances are placed optimally around $H_{\mathbf 1}$ \cite{Huyen2011}.

%In this paper we extend their idea for \mkl\ case and build a metric-learning based \svm\ framework.
%In this paper we use the metric learning view of \svm\ as the foundation for defining several metric-learning-based \svm\ algorithms.

%In this paper we extend the idea presented in \cite{Huyen2011} on the use of the within-class distances and draw a framework for metric-learning-based \svm\ algorithms. We further extend it to the Multiple Kernel Learning framework, which leads to metric-learning-based \mkl\ algorithms.

\cite{Huyen2011} proposed a measure of the within-class distance for \svm.
This measure is inspired by the relation, developed in that paper, between \svm\ and \lmnn---Large Margin Nearest Neighbor \cite{weinberger2009distance}---a popular metric learning algorithm.
It is defined as the sum of the distances of the instances from the margin hyperplane and for the class $C_i$, it is given by:
% It is defined as the total sum of the distances of the instances from the margin hyperplane.
%In \cite{Huyen2011} the authors use the sum of the instance distances to their
%margin hyperplane as a way to control the within-class distance; the respective measure for class $C_i$ is given by:
$d^i_{W_1} = \sum_{{\mathbf x} \in C_i} (|d({\mathbf x}, H_w)| -\gamma) $.
The authors then proposed an \svm\ variant, called \esvm, which optimizes the margin and the above within-class distance measure, essentially combining both the \svm\ and the \lmnn\ learning biases. As we will see below \esvm\ turns out to be a special case of the \svmmetric\ which we will describe in section \ref{subsec:within_band}. The optimization problem of \esvm\ is:
%
% \begin{eqnarray}
% \label{eq:eSVMTotal_org}
% \min_{{\mathbf w},b,\gamma, \vect{\xi}, \vect{\eta}} && \frac{1}{\gamma^2}  +  C \sum_i y_i(\mathbf w^T \mathbf x_i + b) \\
% s.t. &&   y_i({\mathbf w}^T {\mathbf x}_i + b) \geq \gamma  , \,\,\,\, \forall i \nonumber
% \end{eqnarray}
\begin{eqnarray}
\label{eq:eSVMTotal_org}
\min_{{\mathbf w},b}  &&  \mathbf w^T \mathbf w  + \lambda \sum_i max(0,y_i(\mathbf w^T \mathbf x_i + b)-1) \nonumber\\
&& + C \sum_i max(0, 1- y_i(\mathbf w^T \mathbf x_i + b) )
\end{eqnarray}
which is equivalent to:
\begin{eqnarray}
\label{eq:eSVMTotal}
\min_{{\mathbf w},b, \vect{\xi}, \vect{\eta}} &&  {\mathbf w}^T{\mathbf w}  +  C_1\sum_i^n{\xi_i} + C_2 \sum_i^n {\eta_i}  \\
s.t. && 1 - \xi_i \leq y_i({\mathbf w}^T {\mathbf x}_i + b) \leq 1 + \eta_i \nonumber \\
& & \xi_i \geq 0, \eta_i \geq 0, \forall i \nonumber
\end{eqnarray}
% % Its dual form  is:
% % \begin{small}\begin{eqnarray}
% \label{eq:eSVMTotal.dual}\text{Its dual form is:}\,\,\,\, \,\,\,\, \,\,\,\, \,\,\,\, \,\,\,\, \,\,\,\, \,\,\,\, \,\,\,\, \,\,\,\,
%  \max_{\vect{\alpha},\vect{\beta}} & \sum_i^n(\alpha_i - \beta_i) - \frac{1}{2} \sum_{ij}^n (\alpha_i - \beta_i)(\alpha_j - \beta_j) y_iy_j {\mathbf x}_i {\mathbf x}_j \\
%  s.t. & \sum_i(\alpha_i-\beta_i) y_i = 0, \,\,\,\,  0 \leq \alpha_i  \leq C_1; \, 0 \leq \beta_i  \leq C_2, \forall i  \nonumber
% \end{eqnarray}\end{small}
where $\eta_i$ is the distance of the $i$th instance from its margin hyperplane, and $\xi_i$ are the \svm\ slack variables which allow for the soft margin.
This problem is convex and can be kernelized directly as a standard \svm.

\cite{Shivaswamy2010} proposed to maximize the margin and to constrain the outputs of \svm, their optimization problem thus optimizes the margin and some measure of the data spread. This approach also falls within our general metric-learning-based \svm\ framework that we will present right away.

\section{A metric-learning-based \svm\ framework}
\label{sec:emkl}
Since \svm\ can be seen as a metric learning algorithm, we can interpret it in terms of the within- and between-class distances, as is done with typical metric learning algorithms using (\ref{metric.learning.opt}), (\ref{metric.learning.constr}).
The \svm\ margin can be seen as a measure of the between-class distance. Unlike \fda\ where the between-class distance is defined via the distance of the class means, in \svm s, it is defined as the minimum distance of different-class instances from the hyperplane, i.e. twice the \svm\ margin. Unlike other metric learning algorithms, where the between-class distance takes into account \textit{all} pairs of instances of different classes, the \svm\ between-class distance only focuses on pairs of instances of different classes which lie on the margin (in \cite{Huyen2011}, the \svm\ margin is reformulated as: $\min_{i,j, y_i \neq y_j} (\mathbf x_i - \mathbf x_j)^T \mathbf w^T \mathbf w (\mathbf x_i - \mathbf x_j)$). In other words, \svm\ between-class distance, i.e the margin, is a simple example of (\ref{metric.learning.constr}).

Similar to most metric learning algorithms, \svm\ maximizes the \textit{between-class distance}; however, unlike them, it ignores the \textit{within-class distance}. In other words, instances of different classes are pushed far away by the margin, but there is no constraint on instances of the same class.

Interpreting \svm\ as a metric learning problem allows us to fully equip it with the metric learning primal bias, i.e to maximize some between-class distance measure and minimize some within-class distance measure. We propose a learning framework in which in addition to the standard \svm\ margin maximization, we also minimize some measure of the within-class distance. We will call the resulting learning algorithms metric-learning-based \svm. In the next sections we will give different general functions of  the within and between class distances which can be the target of optimization. We will then propose \svm\ specific within-class distance measures, and finally formulated the full learning problem which now will also include some measure of the within class distance.
%Thus, we propose a framework to minimize the within-class distance together with margin maximization.
%We call the resulting \svm s  which optimize both the margin and the within-class distance
%metric-learning-based \svm s. In the next section we show how the within- and between-class distances can be optimized in different ways.

\subsection{Within- and between-class distances cost functions}
\label{sec:optimization}
There are several ways to maximize the between-class distances $ d_{B}$ while minimizing the within-class distances $d_{W}$. Bellow we give some simple and widely used cost functions that include these two terms:
% We introduce some of the cost functions which are simple and widely used: % \cite{Fukunaga90}
\begin{small}
\begin{eqnarray}
\label{ratio.form}
 \max && F_1 =  \frac{d_{B}}{d_{W}} \\
\label{plus.form}
 \max  && F_2 =  d_{B} + \lambda \frac{1}{d_{W}} \\
\label{minus.form}
 \max  && F_3 =  d_{B} - \lambda d_{W} \\
\label{plus2.form}
 \min && F_4 =  \frac{1}{d_{B}} + \lambda d_{W}
\end{eqnarray}
\end{small}
Depending on the exact measures of the within- and between-class distances, these cost functions will lead to convex or non-convex optimization problems. An example of the first cost function $F_1$ of (\ref{ratio.form}) is \fda\ (\ref{fld}) where the resulting optimization problem is convex and easy to solve.
$F_1$ places equal importance on the between- and within-class distances. On the other hand $F_2, F_3$, and $F_4$ allow us to better control their trade-off through the introduction of an additional hyperparameter $\lambda$.
%However, the disadvantage of $F_1$ is it puts the same weight on the importance of the between- and within-class distances. In opposite $F_2, F_3, F_4$ can trade-off the weights of importance, at a cost that they use an extra hyperparameter $\lambda$.
Within our metric-learning-based SVM framework, $d_B$ will denote the margin, and $d_W$ will denote some measure of the within-class distance
%In our metric-learning-based \svm\ framework, $d_B$ is the margin, and $d_W$ can be any measure of the within-class distance.
\subsection{Metric-learning based \svm}
\label{subsec:measures}
In this section we start by presenting a new measure of the within-class distance $d_W$ which is appropriate for \svm. We will then use this measure to define one instantiation of our metric-learning-based \svm\ framework. In addition, we discuss a number of other within-class distance measures which can be used to define other instantiations of our framework.

%In this section we propose a new measure of the within-class distance, $d_{W}$, which is appropriate for \svm. We show how to build a \svm\ variant based on this measure. We also discuss some other proper measures of the within-class distance which can be used within our metric-learning-based \svm\ framework.

\subsubsection{A bandwidth-based within-class distance measure}
\label{subsec:within_band}
One way to control the within-class distance is by forcing the learning instances to stay close to their class margin hyperplane, confining
them within a band defined by their margin hyperplane and a hyperplane parallel to it. The width of this band can be seen as a measure of the within-class distance.
This bandwidth is equal to the maximum distance of the instances to their class margin given by $d^i_{W_2} = \max_{{\mathbf x} \in C_i} |d({\mathbf x}, H_{{\mathbf w}})| - \gamma$. To avoid the effect of outlier instances we add slack variables that allow some of them to lie outside the band.

We introduce now a new \svm\ variant, which optimizes both the margin and the measure of the within-class distance described above. Its cost function trades-off the margin maximization and the
bandwidth minimization. Let $d^i_{W_2} = \epsilon_i \gamma$, $\epsilon_i$ states how many times is the bandwidth of the $C_i$ class
larger than the margin. To simplify the optimization problem we use the same $\epsilon=\epsilon_1=\epsilon_2$ for both bandwidths. Using the cost function given in equation~(\ref{minus.form}), we get the following optimization problem:
\begin{eqnarray}
\max_{{\mathbf w},b,\gamma, \epsilon} && \gamma  - \lambda \epsilon \gamma  \\  %+ C_1 \sum \xi_i + C_2 \sum \eta_i \\
s.t. &&  \frac{y_i({\mathbf w}^T {\mathbf x}_i + b)}{\|\mathbf w\|_2} \geq    \gamma  , \forall i  \nonumber\\
&&  \frac{y_i({\mathbf w}^T {\mathbf x}_i + b)}{\|\mathbf w\|_2} \leq \gamma + \epsilon \gamma , \forall i \nonumber \\
&&  \gamma \|{\mathbf w}\|_2 = 1, \,\, \epsilon \geq 0   \nonumber \\
\Leftrightarrow
\label{eq:eSVM}
\max_{{\mathbf w},b,\epsilon}  && (1 - \lambda\epsilon)/\|{\mathbf w}\|_2^2 \\  %+ C_1 \sum \xi_i + C_2 \sum \eta_i \nonumber\\
s.t. && y_i({\mathbf w}^T {\mathbf x}_i + b) \geq 1 \nonumber\\
&& y_i({\mathbf w}^T {\mathbf x}_i + b) \leq 1+\epsilon , \forall i, \,\, \epsilon \geq 0  \nonumber
\end{eqnarray}
Problem~(\ref{eq:eSVM}) is not convex, however, it can be easily kernelized as standard \svm. If we fix $\epsilon$ then it becomes convex.
Fixing $\epsilon$ is equivalent to fixing a specific value for the margin-bandwidth ratio and then maximizing the margin; this is described by the following optimization problem:
%The scenario with a fixed $\epsilon$ corresponds to the case that with an allowing value of the ratio between margin and the bandwidth, we keep maximizing the margin:
\begin{eqnarray}
\label{eq:eSVM_fixed_e}
% \max_{{\mathbf w},b,\gamma, \vect \xi, \vect \eta} && \gamma + C_1 \sum \xi_i + C_2 \sum \eta_i \nonumber \\
% s.t. && y_i({\mathbf w}^T {\mathbf x}_i + b) \geq \gamma - \xi_i \\
% &&  y_i({\mathbf w}^T {\mathbf x}_i + b) \leq \gamma + \epsilon \gamma + \eta_i, \forall i,\nonumber\\
%  && \gamma \|{\mathbf w}\|_2 = 1, \,\, \epsilon \geq 0 \nonumber\\
% \Leftrightarrow
\min_{{\mathbf w},b,\vect \xi, \vect \eta}&& \|{\mathbf w}\|^2_2 + C_1 \sum \xi_i + C_2 \sum \eta_i \\
s.t. &&  y_i({\mathbf w}^T {\mathbf x}_i + b) \geq 1 - \xi_i   \nonumber\\
&&  y_i({\mathbf w}^T {\mathbf x}_i + b) \leq 1+\epsilon + \eta_i, \forall i \nonumber \\
&& \vect \xi , \vect \eta \geq  0 \nonumber
\end{eqnarray}
where $\vect \xi$ and $\vect \eta$ are the slack variables that allow for the margin and bandwidth violations, the latter
is needed to alleviate the effects of outlier instances.  Intuitively the
hyperparameter $C_2$, which controls the bandwidth slack variable, should be bigger than $C_1$ which
controls the margin slack variable, since we can tolerate more instances outside the band than inside the margin.

%We add the slack variable $\vect \xi$ to have the soft margin and the slack variable $\vect \eta$ to have the soft bandwidth, i.e some instances can be allowed to
%be outside the band, to avoid noisy outliers. Intuitively the hyperparameter $C_2$, which controls the bandwidth slack variable, should be bigger than $C_1$ which
%controls the margin slack variable, since we can tolerate more instances outside the band than inside the margin.
Problem (\ref{eq:eSVM_fixed_e}) is convex, quadratic and
can be solved similarly to standard \svm. We will call it \svmmetric. Its dual form is:
\begin{eqnarray}
\label{dual.form.svmmetric}
 \max_{\vect{\alpha},\vect{\beta}} && \sum_i^n(\alpha_i - (1+\epsilon)\beta_i)  \\
&&  - \frac{1}{2} \sum_{ij}^n (\alpha_i - \beta_i)(\alpha_j - \beta_j) y_iy_j {\mathbf x}^T_i {\mathbf x}_j  \nonumber\\
 s.t. & &\sum_i(\alpha_i-\beta_i) y_i = 0 \nonumber\\
&&  0 \leq \alpha_i  \leq C_1, \forall i \nonumber\\
&& 0 \leq \beta_i  \leq C_2, \forall i   \nonumber
\end{eqnarray}
It can be also kernelized directly as \svm, as the term ${\mathbf x}_i^T \mathbf x_j$ that appears in the dual form (\ref{dual.form.svmmetric}) can be replaced by a kernel function.

Using the $F_4$ cost function (\ref{plus2.form}) we get the following non-convex optimization problem which also ``maximizes the margin while keeping the bandwidth $d^i_{W_1}$ small'':
\begin{eqnarray}
\label{eq:eSVM_ratio}
\min_{{\mathbf w},b,\gamma, \epsilon}  && \frac{1}{\gamma^2} + \lambda\epsilon^2 \gamma^2  \\
s.t. && \frac{y_i({\mathbf w}^T {\mathbf x}_i + b)}{\|\mathbf w\|_2}  \geq \gamma  \nonumber\\
&&  \frac{y_i({\mathbf w}^T {\mathbf x}_i + b)}{\|\mathbf w\|_2} \leq \gamma + \epsilon \gamma, \forall i \nonumber \\
&& \gamma \|{\mathbf w}\|_2 = 1, \,\, \epsilon \geq 0  \nonumber
 \\  \Leftrightarrow
\label{eq:eSVM_ratio_w}
\min_{{\mathbf w},b,\epsilon} && \|{\mathbf w}\|^2_2 + \frac{\lambda \epsilon^2}{\|{\mathbf w}\|^2 _2}  \\
s.t.&&  1 \leq y_i({\mathbf w}^T {\mathbf x}_i + b) \leq 1+\epsilon , \forall i, \,\, \epsilon \geq 0 \nonumber
\end{eqnarray}
However, this problem is non convex even for a fixed $\epsilon$; therefore, we do not explore it further in this paper and we leave it for future work.

%\subsection{Within-class distance measured by the sum of distances from the margin hyperplane}
%\label{subsec:secondmeasure}

Interestingly, we note that if in the optimization problem~(\ref{eq:eSVM}) we set $\epsilon=0$ then this problem reduces to
the optimization problem~(\ref{eq:eSVMTotal}); therefore, we can also solve \esvm\ of \cite{Huyen2011} as a special case of \svmmetric\ using the \svmmetric\ solver.

We note that \svmmetric\ is a general formulation for both \svm\ and \esvm. In the limit when $\epsilon \rightarrow \infty$, \svmmetric\ reduces to standard \svm, and when $\epsilon \rightarrow 0$, \svmmetric\ becomes \esvm.
From the optimization point of view we remark that the bigger value of $\epsilon$ is, the smaller the number of the active constraints will be (see the second set of constraints in (\ref{eq:eSVM_fixed_e})).  Thus in terms of running time, \esvm\ is the slowest, followed by \svmmetric\ and then \svm. However, all are quadratic optimization problems and can be solved efficiently.

% \subsection{Using the \fda\ within-class distance measure}
% An obvious measure of the within-class distance is the one used in \fda, i.e. $d_{W_3} =  {\mathbf w}^T S_{\mathbf w}{\mathbf w}$.
%
% Combining this quantity with the margin maximization we get the following optimization problem:
% \begin{eqnarray}
% \label{eq:eSVMLDA}
% \min_{{\mathbf w},b} &&  {\mathbf w}^T (\lambda S_{\mathbf w} + I ){\mathbf w} \\
% &s.t.& y_i({\mathbf w}^T {\mathbf x}_i + b) \geq 1, \forall i \nonumber
% \end{eqnarray}
% This problem was already introduced in \cite{Tao2005}, however in a very different context to metric learning as we describe here, and without an interpretation of the use of the within and between class distances. We show again this FDA within-class distance measure and its corresponding optimization problem for the reason of completeness of our unified view of the within and between class distance based framework.
%
% (\ref{eq:eSVMLDA}) is convex and can be reformulated as a standard \svm\ problem in a specific transformed space \cite{Tao2005}. However it is not easy to kernelize and \cite{Tao2005} solved it only in the original feature space.

One may also think of the \fda\ within-class distance as an appropriate measure, i.e. $d_{W_3} =  {\mathbf w}^T S_{\mathbf W}{\mathbf w}$. Xiong et al. \cite{Tao2005} combined \svm\ and \fda, optimizing like that the margin and the \fda\ within-class distance.
However, we note that they simply introduced this combination without putting it in the metric learning context that we described here, and they provided no interpretation on the use of the within- and between-class distances.
%However, we note that they just simply introduced this combination without putting it in the metric learning context that we described here.
%without an interpretation of the use of the within and between-class distances, and it was not inspired by the \svm\ metric learning perspective.
%The provided no interpretation on the use of the within and between-class distances, and their approach was not inspired by the \svm\ metric learning perspective.
Still their work falls into our metric-learning-based \svm\ framework as a special case. Using the cost function of (\ref{plus.form}) to optimize the margin and the \fda\ within-class distance, we can formulate the following optimization problem:
\begin{eqnarray}
\label{eq:eSVMLDA}
\min_{{\mathbf w},b} && {\mathbf w}^T (\lambda \mathbf S_{\mathbf W} + I ){\mathbf w} \\
s.t. && y_i({\mathbf w}^T {\mathbf x}_i + b) \geq 1, \forall i \nonumber
\end{eqnarray}
(\ref{eq:eSVMLDA}) is convex and is equivalent to the standard \svm\ problem in the transformed space $ \widetilde{\mathbf x } = \mathbf A \mathbf x = (\mathbf S_{\mathbf W} + \lambda \mathbf I)^{-1} \mathbf x$, where $\mathbf S_{\mathbf W}$ is the \fda\ within-class scatter matrix and  $\mathbf I$ is the identity matrix \cite{Tao2005}. However, it is not straight forward to kernelize and \cite{Tao2005} solved it only in the original feature space.

We note that it is an interesting question to develop more measures for the within- and between-class distances, or the data spread. For example the radius of the smallest sphere containing the data is also a measure of the data spread. Therefore, the different variants of the radius-margin based \svm s \cite{Weston2000,Rakotomamonjy2003,Huyen2009b}, which maximize the margin and minimize the radius, also fall to our metric-learning-based SVM framework.
%If we consider the radius of the smallest sphere containing data is a (loose) measure of the data spread, the radius-margin based \svm s \cite{Weston2000,Rakotomamonjy2003,Huyen2009b}, in which the margin is maximized while the radius is minimized, also fall to our metric-learning-based \svm\ framework.
A more challenging problem is to determine which measure is the best in some specific situations.

\mnote[Removed]{
%
% In this paper we propose a kernelized version of (\ref{eq:eSVMLDA}) as follows.
% We use the representer theorem \cite{Taylor2004}, in which the weights $\mathbf w$ are expressed as a linear combination of the training points,
% i.e. $\mathbf w=\sum_i \alpha_i^n \vect \Phi(\mathbf x_i)$,  following the representer theorem.
% Then the \fda\ within-class distance is computed as follows \cite{Mika99}:
% \begin{eqnarray}
% d^{\Phi}_{W3} = {\mathbf w}S^{\Phi}_{\mathbf w}{\mathbf w} = \vect{\alpha}^T \mathbf N \vect{\alpha}
% \end{eqnarray}
%  where
% $\mathbf N = \sum_{i = 1,2} \mathbf K_i(\mathbf I - \mathbf 1_{n_i}) \mathbf K_i^T$, $\mathbf K_i$ is a $n \times n_i$
% matrix with $(\mathbf K_i)_{lp}:= \mathbf K({\mathbf x}_l, {\mathbf x}^i_p)$ (this is the kernel matrix for class $i$), $\mathbf I$
% is the identity and $\mathbf 1_{n_i}$ the matrix with all entries $1/n_i$.
%
% Problem~(\ref{eq:eSVMLDA}) in the $\vect \Phi(\mathbf x)$ feature space then results in the following convex optimization problem:
% \begin{eqnarray}
% \label{eq:eSVMLDA.kernel}
% \min_{\vect{\alpha},b, \vect{\xi}}  && \vect{\alpha}^T (\mathbf K + \lambda\mathbf N) \vect{\alpha} + C\sum_i^n \xi_i  \nonumber\\
% s.t. && y_i(\sum_j^n \alpha_j \mathbf K({\mathbf x}_j, {\mathbf x}_i) + b) \geq 1 - \xi_i, \xi_i \geq 0, \forall i
% \end{eqnarray}
}
\section{A unified view of \fda, \svm,  metric-learning-based \svm\ and metric learning}
\label{sec:unifiedview}
We first analyze \fda\ from the metric learning perspective as it is done with \svm\ in \cite{Huyen2011}. We focus only on binary classification problems.
%\paragraph{\fda\ from the metric learning perspective}
Similar to \svm, we can also formulate \fda\ as a Mahalanobis metric learning problem, where the transformation matrix is diagonal $\mathbf W$, $diag(\mathbf W) = \mathbf w$. We reformulate the \fda\ between-class, $\widetilde{s_b}$, and within-class, $\widetilde{s_{w}}$, distances with respect to the $H_{\vect 1}$ hyperplane as the standard \fda\ between and within-class distances in the transformed space $\mathbf W \mathbf x$, projected to the norm vector of $H_{\vect 1}$:
\begin{eqnarray}
\widetilde{s_b}  &=&  \frac{1}{d}\| \vect 1^T (\mathbf W {\vect m_i} - \mathbf W{\vect m_j})\|^2 \\
 \widetilde{s_{w}} &= & \frac{1}{d} \sum_{i = 1,2}\sum_{{\mathbf x} \in \mathcal D_i} \|\vect 1^T(\mathbf W {\mathbf x} - \mathbf W \vect{m}_i)\|^2
\end{eqnarray}
The \fda\ learning problem then can be stated in the metric learning jargon as follows: we learn a diagonal transformation $\mathbf W$ so that in the transformed space $\widetilde{\mathbf x} = \mathbf W \mathbf x$, the \fda\ between-class distance, with respect to the $H_1$ hyperplane, is maximized and the \fda\ within-class distances, with respect the same hyperplane, is minimized. This learning problem leads to the standard \fda\ given in (\ref{fld}).
% \begin{wrapfigure}{r}{0.8\textwidth}
\begin{figure}
  \begin{center}
    \includegraphics[scale=0.34]{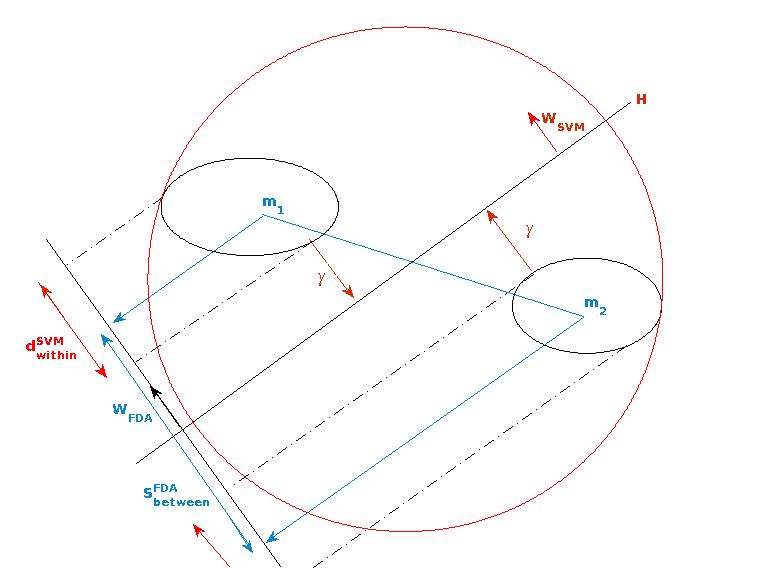}
  \caption{Between and within-class distances for \fda, \svm, and metric-learning-based \svm. While \fda's between-class distance, $s^{FDA}_{between}$, considers the two sample means $\mathbf m_1$ and $\mathbf m_2$, standard \svm\ and metric-learning-based \svm's between-class distance, $d^{SVM}_{between}$, i.e $2\gamma$, considers the two nearest instances of two classes with respect to the hyperplane. The outer sphere is the smallest data enclosing sphere, its radius is a measure of the total data scatter which is minimized by the different radius-margin based \svm\ variants; its minimization together with the margin maximization indirectly minimizes the within-class distances. $d^{SVM}_{within}$ is an example of the measure of the within-class distance of metric-learning-based \svm.}
 \label{fig:svmfda}
  \end{center}
\end{figure}
%\end{wrapfigure}
Thus, we see that from the metric learning perspective both \fda\ and \svm\ learn a diagonal transformation $\mathbf W$. However, the way they define their between and within-class distances is different, although all these distances are based on distances from the fixed hyperplane $H_1$ and are defined \textit{globally}. The \fda\ between-class distance is the squared distance between the two class means, with respect to the hyperplane $H_1: \vect 1^T \mathbf x = 0$, %, i.e  $s_b = \frac{1}{d} \|\vect 1^T \mathbf W (\vect m_i-\vect m_j)\|^2 $,
 while the \svm\ between-class distance is the minimum distance between two instances of the two classes, with respect to $H_1$.
%
% \svm\ interpreted as metric learning problems, learn diagonal transformations and translations that maximize the margin.
% As we already mentioned the margin is a measure of the between-class distance that measures the minimum distance of
% two instances of the two different classes with respect to a hyperplane (Figure 1). From a metric learning perspective  one can argue
% that we should also optimize in addition to the between-class distance, the within-class distance. This argument finds extra support in
% the \svm\ theoretical error bound, which is a function of the radius-margin ratio. Since the radius is a measure of total
% data spread, minimizing this ratio maximizes the margin and minimizes the data spread, thus indirectly minimizing the within
% class distance.
On the other hand, \fda\ defines its within-class distances as the total sum of distances between each instance and its corresponding class mean, while standard \svm\ ignores the within-class distances, i.e there is no constraint on the pairwise distances of the same class instances. Similar to \svm, metric-learning based \svm\ defines its between-class distance as the margin; in addition, it defines its within-class distance in different ways. Radius-margin based \svm s \cite{Weston2000,Rakotomamonjy2003,Huyen2009b}, which are some instantiations of metric-learning based \svm, use the radius of the smallest sphere as a measure of data spread instead of explicitly defining the within-class distance. Figure \ref{fig:svmfda} graphically depicts the difference between standard \svm, metric-learning based \svm, and \fda.

%However as we have shown, the radius is a loose measure of the total data spread, and when we optimize the radius-margin ratio we are in essence optimizing a function of the within and between class distance. Measuring the total data spread via the radius is not consistent with the way the \svm\ between-class distance---the margin, is measured. The radius is computed using the pairwise distances of all instances, while the margin is computed using the distances to a hyperplane. On the other hand, the within-class distance measure proposed in this paper - the bandwidth containing instances, or the one proposed in \cite{Huyen2011} - the total sum distances to the margin hyperplane, are defined consistently with the \svm\ margin, i.e both are based on distances with respect to a hyperplane.
%The proposed metric learning based \svm\ finds support both in the \svm\ theoretical error bound and the learning biases used in metric learning, i.e the preference of large between-class distance and small within-class distance.

The metric-learning-based \svm\ algorithms optimize both the margin and some measures of the within-class distances. Still the constraints on the pairwise distances of the metric-learning-based \svm\ are a simple version of (\ref{metric.learning.constr}). Their within-class distances are defined \textit{globally}, with respect to a hyperplane. Their between-class distance, the margin, only takes into account the closest instances of the two classes.

Metric learning algorithms in general follow the within and between-class distances optimization. However they focus more on details, i.e  pairwise distances are computed \textit{locally}, unlike \fda\ or \svm\ which focus on \textit{global} properties of the pairwise distances using the class means or the instances on the margin.

%In the future it will be a challenging work to build generalization error bound for the presented metric learning problems. To the best of our knowledge there is no such bounds for metric learning. Through the unified view of \svm\ and metric learning we plan to relate the \svm\ error bound with the metric learning problems that we presented in an effort to derive these metric learning bounds.

The derivation of generalization error bounds for the presented metric learning problems is an open issue. To the best of our knowledge no such bounds exist. Through the unified view of metric learning and \svm\ we want to use the error bounds of the latter to derive new bounds for the methods that we present here.

\section{A metric-learning based MKL framework}
\label{sec:MKLinMetricview}
In this section we will show how to exploit the metric-learning approach in the context of \mkl. We first show how to formulate \mklsvm\ as a metric learning problem with the help of two linear transformations. We proceed by developing a metric-learning-based \mkl\ framework which optimizes both the margin and some measure of the within-class distance using multiple kernels.

%In this section we extend further the idea in \cite{Huyen2011} to the \mkl\ scenario. We first show that as \svm, \mklsvm\ too, can be formulated as a metric learning problem with two linear transformations. We then propose a metric-learning-based \mkl\ framework which optimizes both the margin and some measures of the within-class-distances using multiple kernels.
\subsection{MKL from a metric learning perspective}
%In the standard view of \svm, the space is fixed and the hyperplane is moved around to achieve the optimal margin. In the metric view of \svm, the hyperplane is fixed to $H_{\vect 1}$ and the space is scaled, $\mathbf W$, and then translated, $b$, so that the instances are placed optimally around $H_{\vect 1}$.

In \mkl\ we learn linear combinations of kernels of the form $\mathbf{K}_{\vect{\mu}} = \sum_k^m \mu_k\mathbf K_k$. The
feature space $\mathcal{H}_{\vect{\mu}}$ that corresponds to the learned kernel $\mathbf{K}_{\vect{\mu}}$ is given by the mapping:
\begin{eqnarray}
{\mathbf x}  \rightarrow \mathbf \Phi_{\vect{\mu}}({\mathbf x}) =
(\sqrt{\mu_{1}}\mathbf \Phi^T_{1}({\mathbf x}),...,\sqrt{\mu_{m}}\mathbf \Phi^T_{m}({{\mathbf x}}))^T \nonumber\\
\in {\mathcal{H}_{\vect{\mu}}} =  \mathbf{D}_{\sqrt{\vect{\mu}}}\vect{\Phi}\conc({\mathbf x})
\label{eq:MKL.fs}
\end{eqnarray}
where $\mathbf \Phi_k({\mathbf x})$ is the mapping to the $\mathcal{H}_k$ feature space associated with the $K_k$ kernel, $\vect{\Phi}\conc({\mathbf x}) = (\mathbf \Phi^T_{1}({\mathbf x}),...,\mathbf \Phi^T_{m}({{\mathbf x}}))^T \in {\mathcal{H}\conc}$, and $\mathbf{D}_{\sqrt{ \vect \mu}}$ is block diagonal matrix with block diagonal elements ${\sqrt{\vect \mu}}$.

%
%
% We call the feature space we get when $\vect{\mu}$ is set to be $\vect{1}$ the concatenation feature space,
%  $\vect{\Phi}\conc({\mathbf x}) = (\mathbf \Phi^T_{1}({\mathbf x}),...,\mathbf \Phi^T_{M}({{\mathbf x}}))^T \in {\mathcal{H}\conc}$,
% the kernel of which is $\mathbf{K}\conc = \sum_k^M \mathbf{K}_k$.
%
% We denote by $\mathbf{D}_{\sqrt{\mu_k}}$ the diagonal $d_k \times d_k$ matrix which is given by  $\mathbf{D}_{\sqrt{\mu_k}} = \sqrt {\mu_k} \mathbf I$,
% where $\mathbf I$ is the identity matrix, then the feature space induced by the linear combination of kernels  has the following form:
% \begin{eqnarray}
% \label{eq:MKL.matrix.fs}
% {\vect{\Phi}}_{\vect \mu}({\mathbf x}) &=& \left[ \begin{array}{cccc} \mathbf{D}_{\sqrt{\mu_1}} & 0 & 0 & ... \\
% 			                                                  0                & \mathbf{D}_{\sqrt{\mu_2}} & 0 & ...  \\
% 			       0      & 0        &                            & ... \\
% 			       ...    & ...      & .. & \mathbf{D}_{\sqrt{\mu_M}} \\
%  \end{array} \right] \left[\begin{array}{c}  \vect{\Phi}_1({\mathbf x})\\
% \vect{\Phi}_2({\mathbf x})\\
% ...\\
% \vect{\Phi}_M({\mathbf x})\\
% \end{array}\right]  \nonumber\\
% &=& \mathbf{D}_{\sqrt{\vect{\mu}}}\vect{\Phi}\conc({\mathbf x}) \nonumber
% \end{eqnarray}

Similar to \svm, we can also use the fixed hyperplane $H_1$ and view \mklsvm\ as learning a block diagonal linear transformation  $\mathbf{D}_{\sqrt{\vect{\mu}}}$ in the concatenation feature space $\cal H$, followed by
a diagonal linear transformation $\mathbf{W}$ and a translation $b$, such that the margin with respect to the $H_{\vect 1}$ hyperplane
is maximized and the two classes are placed symmetrically in the two different sides of $H_{\vect 1}$.
The linear transformation $\mathbf A$ associated with \mklsvm\ is given by $\mathbf A = \mathbf W  \mathbf{D}_{\sqrt{\vect{\mu}}}$. So \mklsvm\ is also a Mahalanobis metric learning problem where the transformation matrix is  $\mathbf A = \mathbf W  \mathbf{D}_{\sqrt{\vect{\mu}}}$.
\mnote[Removed]{
}

From a metric learning perspective \svm\ uses a single diagonal matrix transformation given by $\mathbf W$
and \mklsvm\ uses two diagonal matrix transformations given by $\mathbf W' = \mathbf W \mathbf D_{\sqrt{\vect{\mu}}}$; both optimize the same cost function (i.e the margin). A formal comparison of the two methods under the metric-learning view can be found in the Appendix.

\subsection{\mkl\ and the optimization of the within-class distance}
Similar to \svm, standard \mklsvm\ optimizes only the margin, i.e the between-class distances but ignores the within-class distances. As before with \svm\ we will now develop a metric-learning-based \mkl\ framework in which we will optimize both the margin and some measure of the within-class distances. We will give two examples of metric-learning-based \mkl\ algorithms, using the \svmmetric, section \ref{subsec:within_band}, and \esvm \cite{Huyen2011}.

%We extend the metric-learning-based \svm\ framework to metric-learning-based \mkl\ framework, in which we optimizes both the margin and a measure of the within-class distances using multiple kernels. Following we show two examples of metric-learning-based algorithms, which are combinations of multiple kernels with \svmmetric\ in section \ref{subsec:within_band} and \esvm\ of \cite{Huyen2011}.

The \svmmetric\ optimization problem, equation~(\ref{eq:eSVM_fixed_e}), in the \mkl\ context becomes:
\begin{eqnarray}
\label{eq:mkl_metric}
    \min_{{\mathbf w},b,\vect{\mu}, \vect{\xi}, \vect{\eta}} && \|{\mathbf w}\|_2^2  +\frac{C_1}{2}\sum_{i}^n \xi_{i}  + \frac{C_2}{2}\sum_{i}^n \eta_{i}   \nonumber\\
s.t. & & 1 -\xi_i \leq y_i(\mathbf{W} \mathbf{D}_{\sqrt{\vect{\mu}}}\mathbf\Phi({\mathbf x}_i) + b) \leq 1+\epsilon  + \eta_i , \forall i \nonumber \\
 && \|\vect{\mu}\|_1 = 1,\mu_k \geq 0, \forall k, \epsilon \geq 0 \nonumber
\end{eqnarray}
%In this section we use these within class distance measures in the context of \mklsvm.
%The optimization problems that use the maximum distance from the margin, eq~(\ref{eq:eSVM}, and the sum of the distances from the margin,
%eq~(\ref{eq:eSVMTotal}), can be directly used within \mklsvm.  For example the simplified version of (\ref{eq:eSVM}), where $\epsilon$ is fixed,
%i.e problem~(\ref{eq:eSVM_fixed_e}) , in the \mkl\ context becomes, (note that here we add the slack variables for soft margin):
%\begin{small}\begin{eqnarray}
%\label{eq:mkl_metric}
    %\min_{{\mathbf w},b,\vect{\mu}, \vect{\xi}, \vect{\eta}} && \|{\mathbf w}\|_p  +\frac{C_1}{2}\sum_{i}^n \xi_{i}  + \frac{C_2}{2}\sum_{i}^n \eta_{i}   \nonumber\\
%s.t. & & 1 -\xi_i \leq y_i(\mathbf{W} \mathbf{D}_{\sqrt{\vect{\mu}}}\vect\Phi({\mathbf x}_i) + b) \leq 1+\epsilon  + \eta_i , \forall i, \,\,\,\, \|\vect{\mu}\|_p = 1,\mu_k \geq 0, \forall k, \epsilon \geq 0 \nonumber
  %\end{eqnarray}
%\end{small}
%which using the $L_1$ norm of $\vect{\mu}$ and the $L_2$ of ${\mathbf w}$  results to:
which is equivalent to:
\mnote[Removed]{
% \begin{small}\begin{eqnarray}
% \label{softMarginMetricMKL1}
% \min_{{{\mathbf w}},b,\epsilon,\vect{\xi},\vect{\mu},\vect{\eta}}& &
% \frac{1}{2} \sum_{k}^{m} \frac{ \langle {\mathbf w}_{k},{\mathbf w}_{k}\rangle}{\mu_{k}}   (1+ \lambda \epsilon) + \frac{C_1}{2}\sum_{i}^n \xi_{i}  + \frac{C_2}{2}\sum_{i}^n \eta_{i}  \nonumber\\ \nonumber
% s.t. &&  1 - \xi_{i} \leq y_{i}(\sum_{k}^{m}\langle {\mathbf w}_{k},  \mathbf \Phi_{k}({\mathbf x}_{i})\rangle + b) \leq 1+\epsilon + \eta_i, \,\,\,\,        \sum_{k=1}^m \mu_{k} = 1, \mu_k \geq 0, \forall k, \xi_i \geq 0, \eta_i \geq 0, \forall i \nonumber
% \end{eqnarray}
% \end{small}
% This is a nonconvex optimization problem. A simplified version of it when $\epsilon$ is fixed is:
}
\begin{eqnarray}
\label{softMarginMetricMKL}
\min_{{\mathbf w},b,\vect{\xi},\vect{\mu},\vect{\eta}}& &
\frac{1}{2} \sum_{k}^{M} \frac{ \langle {\mathbf w}_{k},{\mathbf w}_{k}\rangle}{\mu_{k}}    + \frac{C_1}{2}\sum_{i}^n \xi_{i}  + \frac{C_2}{2}\sum_{i}^n \eta_{i}  \\ \nonumber
s.t. &&  1 - \xi_{i} \leq y_{i}(\sum_{k}^{M}\langle {\mathbf w}_{k},  \mathbf \Phi_{k}({\mathbf x}_{i})\rangle + b)  \leq 1+\epsilon + \eta_i \nonumber\\
&& \sum_{k=1}^M \mu_{k} = 1,  \mu_k \geq 0 , \forall k,  \xi_i \geq 0, \eta_i \geq 0, \forall i \nonumber
\end{eqnarray}
This optimization problem is the counterpart of \svmmetric\ in the \mkl\ context; we will call it \mklmetric. As it was the case with
\svmmetric\ where setting $\epsilon$ to zero leads to \esvm, here too, if in the \mklmetric\ optimization problem~(\ref{softMarginMetricMKL}),
we set $\epsilon=0$ the resulting optimization problem, which we denote by \emkl, will correspond to the coupling of \esvm\ with \mkl.
\emkl\ thus learns kernel combinations that maximize the margin and minimize the within-class distance as the latter is measured by the sum
of the distances from the margin hyperplanes.
% The optimization problem~(\ref{eq:eSVMLDA.kernel}), which makes use of the \fda\ within-class distance measure, can be coupled
% directly with \mkl\ by replacing the kernel matrix with the combination $\sum_k \mu_k \mathbf K_k$; however this replacement
% does not preserve the convexity of eq.~(\ref{eq:eSVMLDA.kernel}).

The \mklmetric\ optimization problem, equation~(\ref{softMarginMetricMKL}), is convex and equivalent to:
\begin{eqnarray}
\label{eq:bachsAlgo} \min_{\vect{\mu}}   J(\vect{\mu}),   \,\,\,\,\,\,  \,\,\,\,\,\,
 s.t.   \,\,\,\,   \sum_{k=1}^m \mu_{k} = 1, \mu_{k} \geq 0, \forall k \
\end{eqnarray}
where:
\begin{small}
 \begin{eqnarray}
\label{insideESVM2}
J(\vect{\mu})  = \left\{
\begin{array}{ll}
 \min_{\mathbf{w},b,\vect{\xi},\vect{\eta}} &
 \frac{1}{2} \sum_{k}^m \frac{\langle \mathbf{w}_{k},\mathbf{w}_{k} \rangle}{\mu_{k}} + \frac{C_1}{2}\sum_{i}^n \xi_{i} + \frac{C_1}{2}\sum_{i}^n \eta_{i}\\
 s.t.                &  1 - \xi_{i} \leq y_{i}(\sum_{k}^{m}\langle \mathbf{w}_{k}, \mathbf{\Phi}_{k}(\mathbf{x}_{i})\rangle + b) \nonumber \\
& \,\,\,\,\,\,\,\,\, \,\,\,\,\,\,\,\,\,\,\,\, \,\,\,\,\,\,\,\,\,\,\,\,\,\,\,\,\,\,\,\,\,\,\,\,    \leq 1 + \epsilon  + \eta_{i}  \nonumber\\
& \forall \xi_i \geq 0, \forall \eta_i \geq 0
\end{array}
	\right.
\end{eqnarray}
\end{small}
We can solve it by a two step algorithm, similar to the one used in SimpleMKL \cite{rakotomamonjy2008}. At the first step of the algorithm we fix $\vect{\mu}$. With fixed $\vect{\mu}$, problem (\ref{softMarginMetricMKL}) becomes a \svm-like optimization problem. In the second step, with the optimal values computed by the \svm-like problem in the first step, we optimize the whole problem by gradient descent with respect to $\vect{\mu}$. The gradient of $J(\vect{\mu})$ is:
$\frac{\partial J(\vect{\mu})}{\partial \mu_{k}} = - \frac{1}{2}  \sum_{ij} (\alpha_i^{*} - \beta_i^{*})(\alpha_j^{*} - \beta_j^{*}) y_i y_j K_{k}(\mathbf x_i, \mathbf x_j)$ where $\alpha^*, \beta^*$ are the solution of the first step.

\mnote[Removed]{
}

\section{Experiments}
\label{sec:expr}
%
% \begin{table}[bt]
% \begin{center}
% \caption{Datasets}
% \label{datasets}
% \vskip 0.15in
%  \scalebox{1}{
% \begin{tabular}{l|lccccc}
%   Datasets                  & \# Sample& \# Feature    \\ \hline \hline
%   Sonar                       & 208     & 60                \\
%   Ionosphere                  & 351     & 34                    \\
%   Musk1                       & 476     & 166             \\
%   Wdbc                        & 569     & 30              \\ \hline
%   CentralNervous              & 60      & 7129                     \\
%   Colon                       & 62      & 2000                     \\
%   Leukemia                    & 72      & 7129                      \\\hline
%   Male vs. Female             & 134     & 1524                     \\
%   Stroke                      & 208     & 2810                       \\
% %  Prostate                    & 322     & 1223                     \\
% %  Ovarian                     & 253     & 771                       \\
%   Liver			      & 354     &6                     \\
%   Wpbc                        & 198      &34         \\
%
% \hline
% \end{tabular}
% }
% \end{center}
% \vskip -0.2in
% \end{table}
We performed two sets of experiments. In the first we examined the performance of three metric-learning-based \svm\ algorithms, namely \svmmetric\ (\ref{eq:eSVM_fixed_e}), \esvm\ (\ref{eq:eSVMTotal_org}) \cite{Huyen2011}, and \svmfda\ (\ref{eq:eSVMLDA}) \cite{Tao2005}. Note that \esvm\ is a special case of \svmmetric\ when $\epsilon$ is set to zero. We used as baseline the performance
of standard \svm\ and \fda. We used the \textit{stprtool} toolbox \cite{stprtool} for \fda.
In the second set, we compared the two metric-learning-based \mkl\ algorithms, i.e.  \mklmetric\ and \emkl, with SimpleMKL \cite{rakotomamonjy2008}, a state-of-the-art \mklsvm\ algorithm.

We experimented with UCI benchmark datasets.
We first standardized the data to a zero mean and one variance. We examine the performance of the different \svm\ based methods over the four following kernels:
linear, polynomial with degree 2 and 3, and a Gaussian kernel with $\sigma = 1$. Kernel $\mathbf{K}$ was normalized as follows: $\mathbf{K}_{ij} = {\mathbf K_{ij}}/\sqrt{\mathbf K_{ii}\mathbf K_{jj}}$. For the \mkl\ methods, we learned combinations of the following 20 kernels:
10 polynomial with degree from one to ten, ten Gaussian with bandwidth $\sigma \in \{0.5, 1, 2, 5, 7, 10, 12, 15, 17, 20\}$,  the same set of basic kernels
as the one used in SimpleMKL~\cite{rakotomamonjy2008}.

Note that \svmfda\  can be only applied on the original feature space since it does not have a kernelized version. For \svmmetric, \esvm, \mklmetric, \emkl\ we set $C_2$ to $C_1/3$, as we can tolerate more bandwidth than margin violations. For \svmmetric\  $\epsilon$ is
set to three, i.e. the allowed band is three times wider than the margin. The  optimal parameter $C$ or $C_1$ of the margin slack variables is chosen by an inner 10-fold cross-validation, from the set of $\mathbf C=\{0.1, 1, 10, 100, 1000\}$.
For \svmfda\ we choose $C$ and $\lambda$ from the same set of  $\{0.1, 1, 10, 100, 1000\}$ using the 10-fold inner cross-validation.
We estimated the classification error using 10-fold cross validation. Folds are the same for all algorithms.

The results for the \svm\ algorithms are given in Table~\ref{tab:svm} and for the \mkl\ ones in Table~\ref{tab:mkl}. We can see that
\svmmetric\ has the lowest error in most of the cases.
Among the 28 different experiments performed for each algorithm (7 datasets $\times$ 4 kernels), \svmmetric\ has the lowest error 19 times,
\esvm\ 10 times, \svm\ 5 times and \fda\ never. We should note that the second constraint of both \esvm\ (\ref{eq:eSVMTotal}) and
\svmmetric\ (\ref{eq:eSVM_fixed_e}) will not be triggered if the data points are already very close to the margin; in such case
\svmmetric\ and \esvm\ have the same performance as \svm.
%In the original space \svmmetric\ is still the best with four wins, followed by \svmfda\ with two wins, \svm\ with one win; \esvm, \fda\ have zero wins.
We also compared the statistical significance of the performance difference using a McNemar's test with the significance level set at 0.05. For
each experiment of a given algorithm, i.e. kernel and dataset, the algorithm was credited with one point if it was significantly better than another algorithm for
the same experiment, half a point if there was no significance difference between the two algorithms, and no point if it was significantly worse than the other
algorithm. Under this ranking schema, in the original feature space, \svmmetric\ got a total score of 20.5 points over the seven datasets, out of a
total possible maximum of 28, followed by \svmfda\ with 17, \svm\ with 15, \esvm\ with 14; \fda\ is by far the worst with only 3.5 points.
In the polynomial kernel space of degree two and three, \svmmetric\ and \esvm\ rank first with 14.5 points but their advantage over \svm\ which
got 13 points is not so pronounced; \fda\ got zero points. With the
gaussian kernel, both \svmmetric\ and \esvm\ have a significant advantage over \svm\ and \fda; the former two
got 14.5 points while \svm\ and \fda\ got only six and seven points respectively. Overall  \svmmetric\ has a consistent advantage
over the different datasets and the different kernels we experimented with.
We should note here that the incorporation of the within-class distance
in the optimization problem seems to bring the largest benefit when the kernel that is used is not that appropriate for the given problem, e.g. the
Gaussian kernel for the datasets we experimented with here, this advantage is not so pronounced when the chosen kernel is good, e.g. polynomial of
degree two and three. In other words incorporating the within-class distance seems to have a corrective effect when a mediocre or poor kernel is used.
 %Nevertheless, \fda\ performs poorly compared to all the others, in all cases.
%Figure \ref{fig:err_linear} shows the cross-validation error of the algorithms in the original feature space, while Figure \ref{fig:err_kernel} shows the results in the kernel spaces.
%  \begin{figure}
%   \begin{center}
%     \includegraphics[scale=0.52]{figures/reportResult.jpg}
%   \caption{Cross validation error in the kernel spaces. Datasets are in the same order as in Table \ref{tab:svm}.
%  \label{fig:err_kernel}}
%   \end{center}
%  \end{figure}
\begin{table*}[t]
\caption{10-fold CV classification error for \svmmetric, \esvm, \svmfda, \svm\ and \fda . Note that \svmfda\ can be applied only in the original feature space, but not in the kernel spaces. A {\bf bold entry} indicates that the respective method had the
lowest classification error.  }
%\vskip 0.15in
\label{tab:svm}
\begin{center}
\scalebox{1}{
\begin{tabular}{|l|c|ccccc|c|c|cccc|}
\hline
Kernel 	&	Datasets	&	\svmmetric 	&	\esvm	&	\svmfda	&	\svm	&	\fda	&	Kernel 	&	Datasets	&	\svmmetric 	&	\esvm	&	\svm	&	\fda	\\
\hline
	%&	colonTumor	&	17.74	&	17.74	&	17.74	&	17.74	&	50	&		&	colonTumor	&	\textbf{24.19}	&	\textbf{24.19}	&	35.48	&	64.52	\\
	%&	centralNervousSystem	&	\textbf{35}	&	\textbf{35}	&	\textbf{35}	&	40	&	\textbf{35}	&		&	centralNervousSystem	&	36.67	&	36.67	&	36.67	&	\textbf{35}	\\
	%&	femaleMale	&	\textbf{5.97}	&	6.72	&	8.21	&	10.45	&	50	&		&	femaleMale	&	\textbf{22.39}	&	\textbf{22.39}	&	29.85	&	50	\\
	%&	Leukemia	&	1.39	&	1.39	&	1.39	&	1.39	&	65.28	&		&	Leukemia	&	\textbf{12.5}	&	\textbf{12.5}	&	33.33	&	65.28	\\
	&	sonar	&	21.63	&	23.08	&	\textbf{20.67}	&	32.69	&	27.88	&		&	sonar	&	13.46	&	\textbf{12.98}	&	13.46	&	26.44	\\
	&	wpbc	&	\textbf{19.7}	&	21.21	&	22.73	&	25.25	&	26.77	&		&	wpbc	&	\textbf{20.2}	&	20.71	&	\textbf{20.2}	&	48.48	\\
	&	ionosphere	&	9.12	&	11.97	&	10.26	&	\textbf{7.98}	&	64.1	&		&	ionosphere	&	\textbf{5.98}	&	6.27	&	6.27	&	30.48	\\
Linear	&	wdbc	&	\textbf{2.11}	&	3.69	&	2.99	&	2.28	&	3.87	& poly3		&	wdbc	&	2.64	&	\textbf{2.28}	&	2.64	&	62.74	\\
	&	liver	&	\textbf{30.43}	&	31.3	&	30.72	&	31.01	&	39.13	&		&	liver	&	\textbf{28.7}	&	30.43	&	29.57	&	40.29	\\
	&	musk1	&	\textbf{13.03}	&	15.76	&	13.24	&	17.02	&	20.8	&		&	musk1	&	\textbf{4.2}	&	4.83	&	\textbf{4.2}	&	36.36	\\
	&	ovarian	&	4.35	&	4.35	&	\textbf{3.95}	&	4.35	&	48.22	&		&	ovarian	&	\textbf{9.49}	&	\textbf{9.49}	&	19.37	&	64.03	\\
\hline	
 Score &McNemar		&	20.5	&	14	&	17	&	15	&	3.5	&Score &McNemar &	14.5	&	14.5	&	13	&	0	\\												
\hline	
\hline								
																									
	%&	colonTumor	&	35.48	&	35.48	&		&	35.48	&	64.52	&		&	colonTumor	&	35.48	&	35.48	&	35.48	&	35.48	\\
	%&	centralNervousSystem	&	35	&	35	&		&	35	&	35	&		&	centralNervousSystem	&	35	&	35	&	35	&	35	\\
	%&	femaleMale	&	\textbf{52.99}	&	53.73	&		&	60.45	&	50	&		&	femaleMale	&	60.45	&	60.45	&	60.45	&	60.45	\\
	%&	Leukemia	&	\textbf{31.94}	&	\textbf{31.94}	&		&	34.72	&	65.28	&		&	Leukemia	&	34.72	&	34.72	&	34.72	&	34.72	\\
	&	sonar	&	17.79	&	\textbf{17.31}	&		&	19.23	&	41.83	&		&	sonar	&	\textbf{34.62}	&	\textbf{34.62}	&	42.79	&	42.79	\\
	&	wpbc	&	\textbf{23.74}	&	\textbf{23.74}	&		&	24.24	&	41.83	&		&	wpbc	&	23.74	&	23.74	&	23.74	&	23.74	\\
	&	ionosphere	&	\textbf{5.41}	&	5.98	&		&	5.98	&	32.76	&		&	ionosphere	&	\textbf{5.41}	&	\textbf{5.41}	&	10.82	&	8.26	\\
Poly2	&	wdbc	&	\textbf{3.51}	&	4.75	&		&	\textbf{3.51}	&	58.52	& gauss1		&	wdbc	&	\textbf{6.5}	&	6.85	&	16.34	&	10.54	\\
	&	liver	&	30.14	&	\textbf{29.86}	&		&	30.72	&	39.98	&		&	liver	&	\textbf{32.17}	&	32.46	&	\textbf{32.17}	&	40.29	\\
	&	musk1	&	\textbf{6.09}	&	6.93	&		&	6.3	&	50.84	&		&	musk1	&	\textbf{39.92}	&	\textbf{39.92}	&	43.49	&	43.07	\\
	&	ovarian	&	\textbf{9.09}	&	\textbf{9.09}	&		&	15.81	&	64.03	&		&	ovarian	&	35.97	&	35.97	&	35.97	&	35.97	\\
\hline
 Score &McNemar   &	14.5	&	14.5	&		&	13	&	0	&		Score &McNemar	&	14.5	&	14.5	&	6	&	7	\\
\hline
%\belowspace
\end{tabular}}
\end{center}
%\vskip -0.1in
\end{table*}

The poor performance of \fda\ can be explained by the between-class distance measure it deploys. Using the class means as a between-class distance measure can lead to class overlapping.
%Moreover focusing only on the ratio of the between and the within-class distances does not really ``maximize the between while minimize the within-class distances'', but only optimize the ratio of them (as discussed in section \ref{sec:optimization}).
\esvm, \svmmetric\ and \svmfda\ do not have this problem since they use the margin as the between-class distance.
Unlike \svmfda\ which can be applied only in the original feature space, since it cannot be kernelized, both \esvm\ and \svmmetric\ can be easily kernelized.
%\esvm\ and \svmmetric\ have advantages over \svmfda\ since they can be easily kernelized (as \svm), unlike \svmfda\ which can be applied only in the original feature space.
\begin{figure}[ht]
%\begin{minipage}[b]{1\linewidth}
\centering
%\subfigure[]{
%\includegraphics[scale=0.35]{figures/SDM2012_time1.jpg}
%%\label{fig:LMNN}
%}
%\subfigure[]{
%\includegraphics[scale=0.35]{figures/SDM2012_time2.jpg}
%%\label{fig:LMNN_SVM}
%}
\includegraphics[scale=0.25]{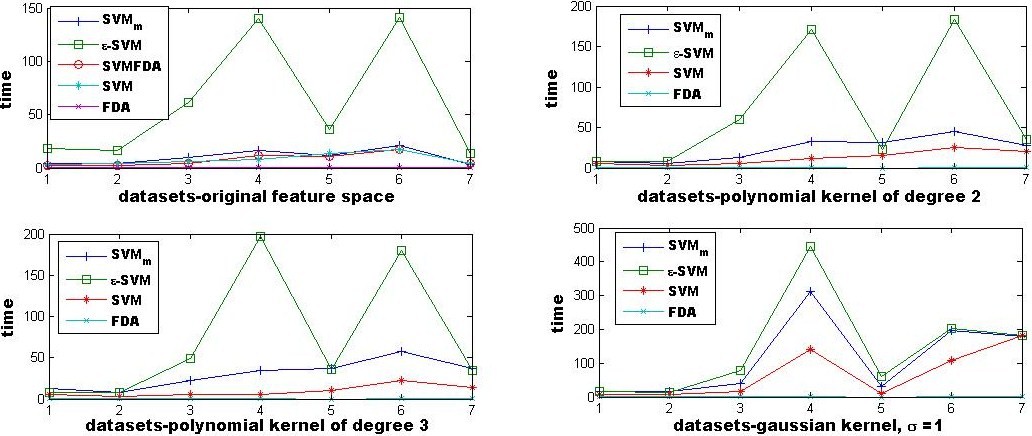}
\caption{Running time of \svmmetric, \esvm, \svmfda, \svm\ and \fda. Datasets are in the same order as in Table \ref{tab:svm}. \label{fig:time}}
\label{fig:views1}
%\end{minipage}
\end{figure}
In Figure \ref{fig:time} we give the running times of the different methods. These are in agreement with our remarks in \ref{subsec:measures}. \esvm\ is the slowest since in general the closer $\epsilon$ is to zero, the more constraints are active and the slower the algorithm will be. \fda\ is the fastest one but as we saw its predictive performance is quite poor.

%an {\em italics entry} indicates that the respective method has an error that is smaller from that of the respective baseline, and an underlined entry indicates that the respective method has an error which is worse than the baseline.

%
% To get a ranking of the three different methods we compared the statistical significance
% of the performance difference using a McNemar's test with the significance level set at 0.05. For each experiment of a given
% algorithm, i.e. kernel and dataset, the
% algorithm was credited with one point if it was significantly better than another algorithm for the same experiment, half
% a point if there was no significance difference between the two algorithms, and no point if it was significantly worse than
% the other algorithm. Thus the total score an algorithm could get for a given experiment was two in the case that it was significantly
% better than both the other algorithms. Under that ranking schema \svmmetric\ got a total score of 53 points over the 48 experiments, out of
% a total possible maximum of 96 points, \esvm\ a score of 51 and \svm\ a score of 40. The strongest advantage of the first two algorithms
% appears in the Gaussian and third degree polynomial kernel.
%
%%%%%%% MKL results
\mnote[Remove]{
\color{red}
We do the comparison on the some UCI datasets and benchmark datasets used in~\cite{Shivaswamy2010} and \cite{Gunnar2001}. We used the same 100 random splits as in these two papers,  For UCI datasets, we generated 100 splits following the same rules, i.e 70\% data of each split is used for training and 30\% is used for testing. We tested the statistical significance by t-test at 5\% significance level. The scoring schema is the same as in McNemar's test.

\begin{table}[ht]
 \begin{center}
\caption{Average classification error for  SimpleMKL, \mklmetric\ and \emkl. A {\bf bold entry} indicates that the respective method had the lowest classification error. An \textit{italic entry} indicates that the respective methods had a lower error than SimpleMKL }
\vskip 0.1in
\label{tab:mkl}
 \scalebox{0.84}{
\begin{tabular}{|l|cc|ccc|}
\hline
 Datasets      	&  \# Instances & \# Features & SimpleMKL 	&\mklmetric  &\emkl 	\\
\hline
 Sonar 		&208	&60		& 16.35		& \textit{13.46}		   & \textbf{12.5} \\
 Ionosphere  	&351	&34		& 4.27		& \textbf{3.70}		   &\textit{3.99}\\
 Musk1      	&476	&166		& 5.25          & \textit{4.41} 		   &\textbf{4.20}\\
 Wdbc 		&569	&30		& 2.64		&\textbf{ 2.11}		   &2.64\\
 %CentralNervous & 35.00		& 35.00		   & 35.00 \\
 %Colon   	 & 38.71		& \textbf{20.97}   &\textbf{20.97}\\
 %Leukemia    	& \textbf{1.39}       &\textbf{1.39}	   & \underline{2.78}\\
 %MaleVSfemale   & 11.19	        &11.19             &\textbf{8.21}\\
 Stroke    	&208	&2810		& 23.08		& \textit{22.60}		   &\textbf{22.12}\\
% Prostate      &21.43		& 21.43		&21.43		   &21.43\\
 Ovarian    	&253	&771		& 4.35		& 4.35		   	&\textbf{3.95}\\
 Liver    	&354	&6		& 34.20		& 34.49			&\textbf{32.46}\\
 Wpbc    	&198	&34		& 23.23		& \textit{22.73	}	&\textbf{22.73}\\

\hline
Score t-test   &  	&		&7.5		& 8.5      		&    8   \\
\hline
\end{tabular}
}
 \end{center}
\vskip -0.15in
% \end{table}
\end{table}

\color{black}
%%%%
}

In Table ~\ref{tab:mkl} we give the results for the \mkl\ experiments. \emkl\ and \mklmetric\  have now only a slight advantage compared to
SimpleMKL. \emkl\ has the lowest error in six datasets while \mklmetric\ is the best in two datasets. In terms of the McNemar score, \mklmetric\
got 8.5 points, followed by \emkl\ with eight points and SimpleMKL with 7.5 points. Unlike standard \svm\ the incorporation of the within-class
distance in the \mkl\ cost function does not seem to deliver a significant performance improvement. This is somehow in agreement with our previous
observation, i.e. that the incorporation of the within-class distance measure seems to have a strong positive effect when the kernel that is used
is not appropriate. By learning the kernel as \mkl\ does we overcome the problem of a possible poor kernel selection, provided that within
the kernel set over which we learn there are appropriate kernels for the given problem.

\begin{table}[ht]
 \begin{center}
\caption{10-fold CV classification error for  SimpleMKL, \mklmetric\ and \emkl. A {\bf bold entry} indicates that the respective method had the lowest classification error. An \textit{italic entry} indicates that the respective methods had a lower error than SimpleMKL }
\vskip 0.1in
\label{tab:mkl}
 \scalebox{0.84}{
\begin{tabular}{|l|cc|ccc|}
\hline
 Datasets      	&  \# Instances & \# Features & SimpleMKL 	&\mklmetric  &\emkl 	\\
\hline
 Sonar 		&208	&60			& 16.35		& \textit{13.46}		   & \textbf{12.5} \\
 Ionosphere  	&351	&34		& 4.27		& \textbf{3.70}		   &\textit{3.99}\\
 Musk1      	&476	&166		& 5.25          & \textit{4.41} 		   &\textbf{4.20}\\
 Wdbc 		&569	&30		& 2.64		&\textbf{ 2.11}		   &2.64\\
 %CentralNervous & 35.00		& 35.00		   & 35.00 \\
 %Colon   	 & 38.71		& \textbf{20.97}   &\textbf{20.97}\\
 %Leukemia    	& \textbf{1.39}       &\textbf{1.39}	   & \underline{2.78}\\
 %MaleVSfemale   & 11.19	        &11.19             &\textbf{8.21}\\
 Stroke    	&208	&2810		& 23.08		& \textit{22.60}		   &\textbf{22.12}\\
% Prostate      &21.43		& 21.43		&21.43		   &21.43\\
 Ovarian    	&253	&771		& 4.35		& 4.35		   	&\textbf{3.95}\\
 Liver    	&354	&6		& 34.20		& 34.49			&\textbf{32.46}\\
 Wpbc    	&198	&34		& 23.23		& \textit{22.73	}	&\textbf{22.73}\\
\hline
Score McNemar   &  	&		&7.5		& 8.5      		&    8   \\
\hline
\end{tabular}
}
 \end{center}
\vskip -0.15in
% \end{table}
\end{table}
From the results, it is apparent that one does not need only to control for the between-class distance, as standard \svm\ does, but also for the within-class distance, since the incorporation of some measure of the latter in the optimization problem considerably improves the performance over the standard \svm. When it comes to determining which within-class distance measure is more appropriate then the width of the band containing the instances that we deployed in \svmmetric\ has a clear advantage since it does not only lead to the best performance but it results in a convex optimization problem which is easy to kernelize and in addition includes as a special case \esvm.

%
% Concerning the question of which measures are the best for the between and within-class distances, we can answer empirically that the margin, i.e the minimum distance of instances of two classes is better than using the class means. Certainly we should have a mechanism to include also the soft version of between-class distances (e.g by using slack variables). For the within-class distance, the \svmmetric's measure shows more advantage since it leads to a convex optimization problem which also includes \esvm\ as a special case, and is easier to solve and to kernelize than the \svmfda's. Moreover \svmmetric\ shows better performance comparing to both \svmfda\ and \esvm.

\section{Conclusion}
\label{sec:conclusion}
Inspired by recent work that investigated the relations of \svm\ and metric learning \cite{Huyen2011}, we present here a novel framework that equips \svm, as well as \mkl, with metric learning concepts. The new algorithms that we propose optimize not only the standard \svm\ margin, which can be seen as a measure of the between-class distance, but in addition they also optimize measures of the within-class distance. For the latter we propose a new measure, the width of the band along the margin hyperplane that contains the learning instances, and we derive new \svm\ and \mkl\ variants that exploit. Our experimental results show that we achieve important predictive performance improvements if we include measures of the within-class distance in the optimization problem. In addition the new algorithms that we derived are convex and easy to kernelize.

%
% We develop a metric-learning-based \svm\ and \mkl\ framework in which we equip \svm, \mkl\ with metric learning ideas.
% The algorithms proposed under these framework optimize both the \svm\ between-class distance, i.e the margin, and different measures of the within-class distances using a single kernel or multiple kernels. We propose a new measure of the within-class distance which is the bandwidth containing the instances along the margin hyperplane, and we derive new variants of \svm, \mkl\ which optimizes both the margin and the above within-class distance.
% The experimental results show that the inclusion of the within-class distance
% minimization improves the classification performance, compared to that of standard \svm\ and \mkl.

There are a number of additional issues we plan to examine.
The most challenging one is the derivation of generalization error bounds for the presented metric learning
problems. To the best of our knowledge no such bounds exist for metric learning. Through the unified view of
\svm\ and metric learning we want to relate the \svm\ error bound with the presented metric learning problems. It is also a challenge to determine which measures of between and within-class distances are the best for a specific problem.
Additionally, and similar to \cite{Bach2005} and \cite{Efron2004}, we want to solve for the full regularization path, for $\epsilon$ or $\lambda$.
%Finally we plan to extend the metric-learning-based \svm\ framework for multiple kernel learning.
%\begin{small}

\section*{Appendix}
We here present a formal comparison of \svm\ and \mklsvm\  under the metric-learning view.

%\subsection*{Norm constraints in \svm\ and \mklsvm}
\label{sec:onTheNorms}
%Before proceeding we will give a brief discussion on how \svm\ and \mklsvm\ control the scaling of their learned linear transformations, and what is the relation between these two ways of controlling the scaling.
From a metric learning perspective \svm\ uses a single diagonal matrix transformation given by $\mathbf W$
and \mklsvm\ a two diagonal matrix transformation given by $\mathbf W' = \mathbf W \mathbf D_{\sqrt{\vect{\mu}}}$; both optimize the same cost function (i.e the margin).
%Note that we can use the same type of decomposition into two matrices and define the respective optimization problem also for standard \svm.
%\footnote{Such a two diagonal matrices transformation has already been used in the context {\em R-}\svm, ~\cite{Huyen2009b}, Chapter \ref{chap:RSVM}, an extension of standard \svm\ which instead of optimizing only the margin optimizes a convex approximation of the radius-margin ratio}.
We will compare \svm\ and \mklsvm\ by comparing the use of one and two diagonal transformation matrices.

For \mklsvm\ the $\mathbf D_{\sqrt{\vect{\mu}}}$ matrix is block diagonal,
while if we use that decomposition with \svm\ it is a simple diagonal matrix.
%To avoid the problem of scaling of the margin with the uniform scaling of the $diag({\mathbf W})$ or the $diag(\mathbf{W}')$ and $diag(\mathbf D_{\sqrt{\vect{\mu}}})$ vectors, we control their norm.
%To simplify the discussion we will take the case of standard \svm\ and
We denote by $\mathbf{w}=diag({\mathbf W})$, $\mathbf{w}'=diag(\mathbf{W}')$
and $\sqrt{\vect{\mu}}=diag(\mathbf D_{\sqrt{\vect{\mu}}})$.
%To control the scaling of the margin when we use the standard formulation with a single vector we set  $\|\mathbf{w}\|_p = 1$, and if we use the two vectors decomposition we set $\|\mathbf{w'}\|_p = 1$ and $\|\vect{\mu}\|_p = 1$ (in fact this is what \mklsvm\ does).
The optimization problems of \svm\ and \mklsvm\ can be formulated as follows:
%Then we have the following two optimization problems:
\begin{eqnarray}
\label{eq:normp}
\text{SVM:} \max_{\mathbf{w},b} && \gamma \\
s.t & & y_i(\mathbf{w}^T \mathbf{x}_i + b) \geq \gamma, \forall i \nonumber \\
&& \|\mathbf{w}\|_p = 1 \nonumber
\end{eqnarray}

\begin{eqnarray}
\label{eq:norm2}
\text{MKL:} \max_{\mathbf{w'},b} && \gamma \\
s.t & & y_i((\mathbf{w'} \circ \vect{\mu})^T  \mathbf{x}_i + b) \geq \gamma, \forall i \nonumber \\
&& \|\mathbf{w'}\|_p = 1, \,\,  \|\vect{\mu}\|_p = 1, \mu_i \geq 0 \nonumber
\end{eqnarray}

We now try to highlight relationship of these two optimization problems.
%The operator $\circ$ denotes the elementwise product of two vectors.
Given some $\vect \mu$ vector such that $\mu_i \neq 0, \forall i$, $\|\vect{\mu}\|_p=1$, and $\mathbf w= \mathbf{w'} \circ \vect{\mu}$,  we define a new norm $l_{\vect \mu}(\mathbf w)$ as follows: $l_{\vect \mu}(\mathbf w)=\|\mathbf{w'}\|_p$.
% \begin{eqnarray}
% \label{newnorm}
% l_{\vect \mu}(\mathbf w)=\|\mathbf{w'}\|_p
% \end{eqnarray}
It is easy to prove that $l_{\vect \mu}({\mathbf w})$ satisfies all the properties of a norm, which means it is a valid norm.

Using this new norm, we can rewrite problem~(\ref{eq:norm2}) as:
\begin{eqnarray}
\label{eq:norm3}
\max_{\mathbf{w},b} \,\,\, \gamma &&\,\,\,\,\,\,
s.t. \,\,\,\,\, y_i(\mathbf{w}^T  \mathbf{x}_i + b) \geq \gamma, \forall i  \\
&& \,\,\,\,\,\,\,l_{\vect \mu}(\mathbf{w}) = 1, \,\,  \|\vect{\mu}\|_p = 1 \nonumber
\end{eqnarray}
Let $T_{\mathbf{w}}$ be the feasible set of $\|\mathbf{w}\|_p = 1$, obviously
$T_{\mathbf{w}}$ is also the feasible set  $\|\mathbf{w'}\|_p = 1$; $T_{\vect{\mu}}$ is the feasible set of $\|\vect{\mu}\|_p = 1$.
Let $T_{\mathbf{w'} \circ \vect{\mu}}=\{\mathbf{w'} \circ \vect{\mu}| \vect{\mu} \in T_{\vect{\mu}}, \mathbf{w'}\in T_{\mathbf{w}}\}$.
For a given value of  $\vect \mu= \vect c$ there is a one-to-one mapping from
$T_{\mathbf{w}}$ to $T_{\mathbf{w}' \circ \vect \mu=\vect c}$. The cardinality of the $T_{\mathbf w}$ feasible set
is much smaller than the cardinality of $T_{\mathbf{w}' \circ \vect{\mu}}$ therefore using this new norm gives
more flexibility in finding a solution of the optimization problem, which could potentially lead to a better
solution.

\begin{figure}
\begin{center}
\includegraphics[scale=0.4]{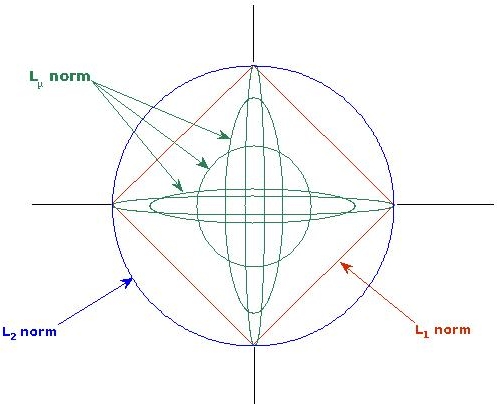}
\end{center}
\caption{Relation of $L_1$, $L_2$ and $L_{\vect \mu}$ norms} \label{fig:norms}
\end{figure}

In Figure~\ref{fig:norms} we illustrate the relations of the different norms just discussed in the two
dimensional space.  We compare the p=1 and p=2 norms of $\mathbf w$ to its $l_{\vect \mu}$ norm.
We fix $\|\mathbf w\|_1 =1$, its feasible set $T_{\mathbf{w}}$ is the diamond, $\|\mathbf w\|_2 =1$, its feasible set $T_{\mathbf{w}}$ is the outer
circle, and $l_{\vect \mu} (\mathbf w) =1$, $\|\vect \mu \|_1=1$, for which we have the set of feasible sets, $T_{\mathbf{w}' \circ \vect{\mu}}$,
that correspond to the different values of $\vect \mu$ that satisfy $\|\vect \mu\|_1=1$, which are given
by the inner ellipsoids.

\bibliography{ECML2012}
\bibliographystyle{splncs03}

% that's all folks
\end{document}